\documentclass{ecai} 

\usepackage{latexsym}
\usepackage{amssymb}
\usepackage{amsmath}
\usepackage{amsthm}
\usepackage{booktabs}
\usepackage{enumitem}
\usepackage{graphicx}
\usepackage{color}
\usepackage{algorithm}
\usepackage{algorithmic}
\usepackage{colortbl}
\usepackage{svg}
\usepackage{multirow}
\usepackage{subcaption}
\usepackage{caption}
\usepackage{booktabs} 
\usepackage{subfig}

\newcommand{\BibTeX}{B\kern-.05em{\sc i\kern-.025em b}\kern-.08em\TeX}

\begin{document}


\begin{frontmatter}


\paperid{2613} 


\title{Feature Bank Enhancement for \\ Distance-based Out-of-Distribution Detection}


\author[1]{\fnms{Yuhang}~\snm{Liu}}
\author[1]{\fnms{Yuefei}~\snm{Wu}}
\author[1]{\fnms{Bin}~\snm{Shi}\thanks{Corresponding Author. Email: shibin@xjtu.edu.cn}} 
\author[1]{\fnms{Bo}~\snm{Dong}} 

\address[1]{School of Computer Science and Technology, Xi'an Jiaotong University}


\begin{abstract}
Out-of-distribution (OOD) detection is critical to ensuring the reliability of deep learning applications and has attracted significant attention in recent years. A rich body of literature has emerged to develop efficient score functions that assign high scores to in-distribution (ID) samples and low scores to OOD samples, thereby helping distinguish OOD samples. Among these methods, distance-based score functions are widely used because of their efficiency and ease of use. However, deep learning often leads to a biased distribution of data features, and extreme features are inevitable. These extreme features make the distance-based methods tend to assign too low scores to ID samples. This limits the OOD detection capabilities of such methods. To address this issue, we propose a simple yet effective method, Feature Bank Enhancement (FBE), that uses statistical characteristics from dataset to identify and constrain extreme features to the separation boundaries, therapy making the distance between samples inside and outside the distribution farther. We conducted experiments on large-scale ImageNet-1k and CIFAR-10 respectively, and the results show that our method achieves state-of-the-art performance on both benchmark. Additionally, theoretical analysis and supplementary experiments are conducted to provide more insights into our method.  
\end{abstract}

\end{frontmatter}


\section{Introduction}

Most deep learning models are trained under the closed-world assumption, where all test data are assumed to follow the same distribution as the training data, known as in-distribution (ID) samples. However, this assumption does not always hold in real-world scenarios \citep{drummond2006open}. The deployed models will inevitably encounter unseen examples that deviate from the training distribution, known as out-of-distribution (OOD) samples. Unfortunately, these OOD samples can cause deployed models to make over-confidence and harmful predictions \citep{nguyen2015deep}. For example, a medical diagnostic system may fail to recognize diseases that did not appear in the training set, potentially leading to misdiagnosis. If the system identifies unrecognized diseases as OOD and warns doctors in advance, such medical accidents can be prevented. Therefore, detecting and handling these OOD inputs can be paramount in high-risk applications such as autonomous driving \citep{huang2020survey} and medical diagnostics \citep{litjens2017survey}. 

\begin{figure}[ht]
    \centering 
    \begin{subfigure}[b]{0.225\textwidth}
        \centering
        \includegraphics[width=\textwidth]{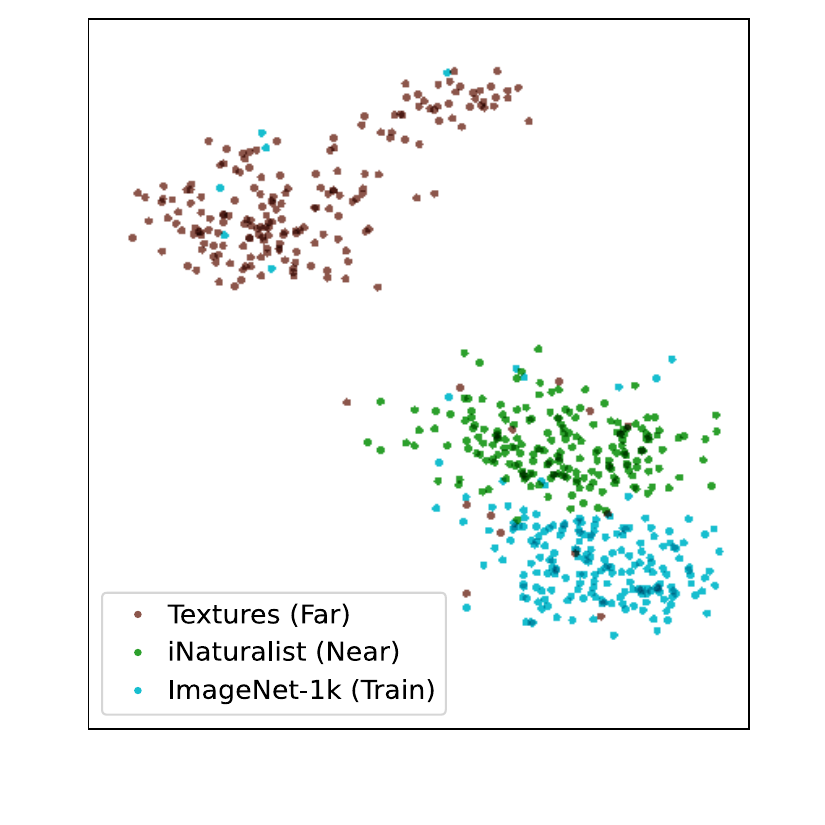}
        \caption{t-SNE Visualization}
        \label{fig1a}
    \end{subfigure}
    \hspace{0.01\textwidth}
    \begin{subfigure}[b]{0.225\textwidth}
        \centering
        \includegraphics[width=\textwidth]{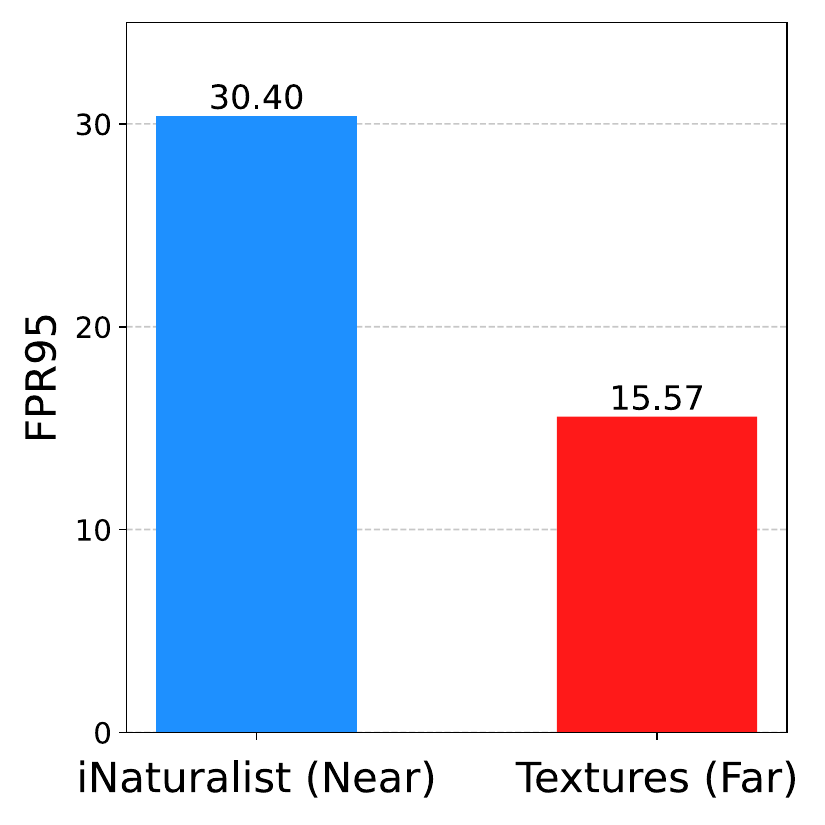}
        \caption{OOD Detection Results}
        \label{fig1b}
    \end{subfigure}
    
    \caption{\textbf{Left: t-SNE visualization.} We use t-SNE~\protect\citep{van2008visualizing} to visualize the feature embeddings of the training data (ImageNet-1k~\protect\citep{deng2009imagenet}), near-OOD samples (iNaturalist~\protect\citep{van2018inaturalist}) and far-OOD samples (Textures~\protect\citep{cimpoi2014describing}), which are defined in OpenOOD~\protect\citep{yang2022openood}. The features are extracted from the penultimate layer of ResNet-50~\protect\citep{he2016deep} model trained on ImageNet-1k. \textbf{Right: Detection performance.} The detection performance of KNN~\protect\citep{sun2022out} on these two kinds of OOD datasets under FPR95 metric. Lower FPR95 value corresponds to better performance.}
    \label{fig1}

\end{figure}

Recently, numerous methods for OOD detection have been proposed \citep{yang2021generalized}. Among them, distance-based methods are widely used because they are simple but effective, and do not require additional training \citep{lee2018simple,sehwag2021ssd,sun2022out,park2023nearest}. The core idea of these methods is to leverage the feature extracted by deep learning models and measure the similarity or dissimilarity between the inputs and the training sample by calculating the distance. For instance, KNN \citep{sun2022out} explores the efficacy of non-parametric nearest-neighbor distance for OOD detection, and ViM \citep{wang2022vim} utilizes information from the feature space to improve the basic score function. However, distance-based methods may struggle with fine-grained detection performance in certain kinds of OOD data.


Distance-based methods typically rely on the assumption: OOD sample features are relatively distant from the training sample features \citep{sun2022out}. When OOD samples contain obvious domain shifts from the training data (\emph{e.g.}, OOD samples from a different domain), they are located far from the training data in the feature space (named as far-OOD samples) \citep{yang2022openood}. For far-OOD samples, distance-based methods can detect them without difficulty. However, when OOD samples exhibit only semantic shifts (\emph{e.g.}, OOD samples are drawn from the same domain but lie outside the training label space), they tend to remain close to the training data in the feature space (named as near-OOD samples) \citep{yang2022openood}. This causes distance-based methods to mistakenly assign high scores to these near-OOD samples, similar to those of ID samples, resulting in poor detection performance on near-OOD samples. In Figure \ref{fig1}, we use t-SNE \citep{van2008visualizing} to visualize the feature embeddings of the training data and two kinds of OOD samples, and demonstrate the performance of one distance-based method (KNN \citep{sun2022out}) on them. These results demonstrates that while distance-based methods perform well on far-OOD samples, their effectiveness drops on near-OOD samples.

\begin{figure}[t!]
    \centering

    \begin{subfigure}[b]{0.22\textwidth}
        \centering
        \includegraphics[width=\textwidth]{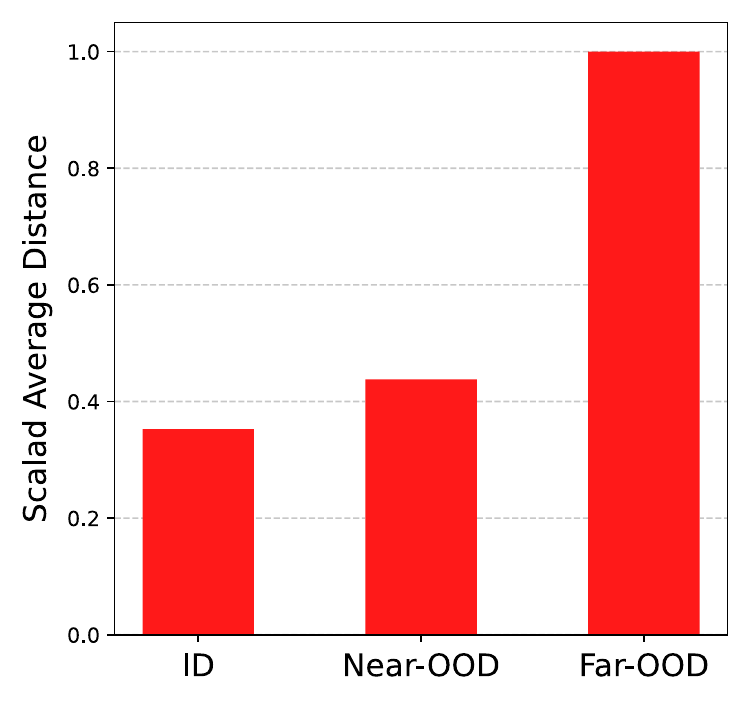}
        \caption{\textbf{Typical} training features}
        \label{fig2a}
    \end{subfigure}
    \hspace{0.03\textwidth}
    \begin{subfigure}[b]{0.22\textwidth}
        \centering
        \includegraphics[width=\textwidth]{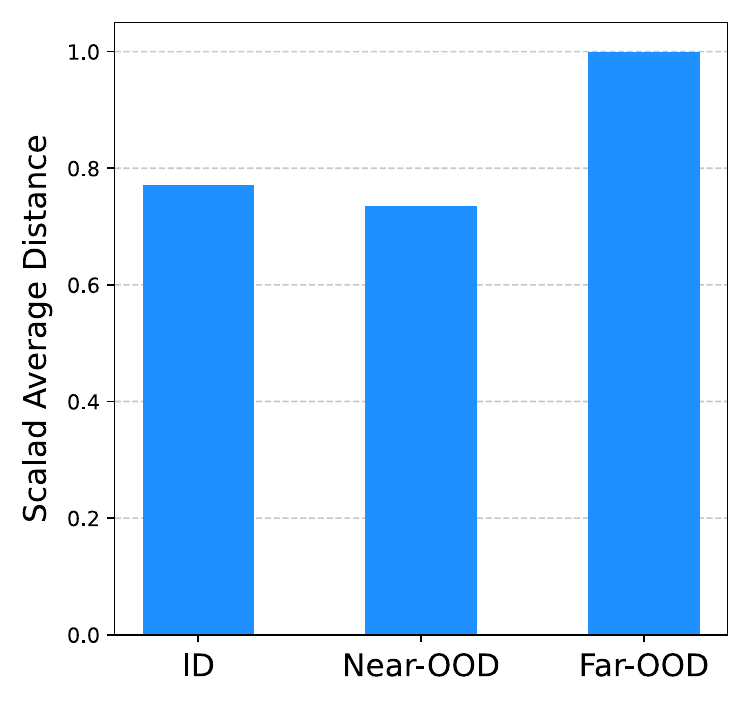}
        \caption{\textbf{Extreme} training features}
        \label{fig2b}
    \end{subfigure}
    \caption{\textbf{Scaled Average distances.} We compute the distances from two kinds of training features to ID, near-OOD and far-OOD features in one feature channel, and present their mean values. 
    The results are scaled by normalizing the maximum value to 1 for more intuitive visualization and comparison.}
    \label{fig2}
\end{figure}

Previous distance-based methods overlook the intrinsic defects caused the by outlier features within the training data. He \emph{et al}. revealed that outlier features are commonly observed in deep learning models \citep{he2024understanding}. As shown in Figure \ref{fig1a}, we also observed that while most training features are clustered near the center, some are distant from the center and close to OOD features. We divide the training features into typical training features (close to the center) and extreme training features (distant from the center). 
Ideally, the distances from training features, whether typical or extreme training features they are, to ID features should be smaller than those to OOD features. 
In Figure \ref{fig2}, we illustrate the distances from these two kinds of training features to ID, near-OOD, and far-OOD features. Figure \ref{fig2a} shows that typical training features follow the distance-based methods' assumption, but Figure \ref{fig2b} reveals that the distances from extreme training features to ID features are larger than those to near-OOD features, which contradicts the basic assumption. 
When calculating OOD scores, these extreme training features tend to assign high scores to near-OOD samples, which might even exceed those assigned to ID samples.
Given this potential negative impact, properly dealing with these extreme training features is a key to improve the detection capability of distance-based methods.

In this paper, we propose a simple yet effective method, \textbf{F}eature \textbf{B}ank \textbf{E}nhancement (\textbf{FBE}), to improve distance-based OOD detection methods. Specifically, we utilize statistical measure from the training feature to determine which features are significantly deviated from the center, and then we constrain these extreme features to the separation boundaries. Consequently, the distances from these extreme training features to most ID features will diminish, and distances to near-OOD features will increase. As a result, compared to the original distance-based methods, ID samples receive higher scores, and near-OOD samples receive lower scores. In this way, our method not only enhances performance of distance-based methods for near-OOD samples, but also maintains their original detection capabilities on far-OOD samples. Additionally, since it does not require any modifications to the training processes, our method can be conveniently implemented on a pre-trained model, without impacting the model’s primary task (such as classification). We conduct an extensive evaluation on the widely used ImageNet-1k \citep{deng2009imagenet} and CIFAR-10 \citep{krizhevsky2009learning} benchmarks, achieving state-of-the-art performance. 
We also provide theoretical analysis and supplementary experiments to explain the working mechanism of FBE. 
The main contributions of our paper are summarized as follows:

\begin{enumerate}
    \item We propose \textbf{F}eature \textbf{B}ank \textbf{E}nhancement (\textbf{FBE}), a simple yet effective approach that mitigates the negative impact of extreme training features on distance-based OOD detection methods. FBE is easy to implement and does not require changes to existing model architectures.
    \item We conduct a comprehensive evaluation of FBE across various OOD benchmarks, establishing a \textbf{state-of-the-art} performance among all types of OOD score functions.
    FBE can improve the fine-grained detection capabilities of distance-based methods for near-OOD samples, while also maintaining their good detection performance on far-OOD samples.
    Moreover, FBE can further enhance the performance of existing network truncation methods.
    \item We provide theoretical analysis and conduct supplementary experiments to explain the working mechanism of FBE, thereby underpinning the superior empirical results and improving the understanding of our approach.
\end{enumerate}

\section{Related Works}
Nguyen \emph{et al}. first revealed the phenomenon of neural networks' overconfidence in OOD data \citep{nguyen2015deep}. This discovery has since gained increasing research interest on OOD detection. 
Among all OOD detection methods, \emph{post-hoc} methods, which utilize pre-trained models without the need for retraining, offer significant advantages in real-world scenarios.
Post-hoc OOD detection research mainly falls into two categories: network truncation and designing score functions.

\subsection{Network truncation}
Such methods modify specific layers in the network to increase the distinction between ID and OOD output signals.
For example, methods like ReAct, RankFeat and BATS employ various clipping techniques on the output signals of the hidden layers.
ReAct \citep{sun2021react} uses truncated activation to eliminate abnormal activations caused by OOD samples.
RankFeat \citep{song2022rankfeat} subtracts the primary singular matrix of the feature map in deep layers.
BATS \citep{zhu2022boosting} curtails the output of batch normalization layer into a compact interval.
Besides clipping techniques, ODIN \citep{liang2017enhancing} introduces gradient-based input perturbations and employs temperature scaling.
However, these model modification approaches cannot be used independently and require integration with a score function to realize their full potential in OOD detection. Additionally, they require specific modifications to the network architecture, which limits their generalizability across different models.

\subsection{Designing score function}
This kind of methods analyze the characteristics of model outputs and devise score functions with discriminative capabilities, which assign significantly different scores to ID and OOD samples.
Building upon this conceptual framework, these methods can be further classified into classifier-based and distance-based methods \citep{yang2022openood}.

\textbf{Classifier-based methods} utilize logits from the classification layer, which are related to confidence, to calculate the OOD score.
MSP \citep{hendrycks2016baseline}, the most fundamental baseline in this field, uses maximum SoftMax as the score function. 
Energy \citep{liu2020energy} examines the limitation of the SoftMax function and propose to use energy function.
GradNorm \citep{huang2021importance} focuses on gradient statistics, with the aims of minimizing the KL score.
Considering the information loss occurs during the transition from the hidden layer to the classification layer, a notable limitation of classifier-based methods is that they have not fully utilized the outputs of the hidden layers, which contain valuable information for OOD detection.

\textbf{Distance-based methods} typically calculate the distances in high-dimensional feature spaces, which contain rich semantic information, to differentiate between ID and OOD samples. SSD \citep{sehwag2021ssd} introduces a self-supervised representation learning approach and weakens the assumption of multivariate Gaussian distributions, making it more suitable for real-world scenarios.
KNN \citep{sun2022out} is a non-parametric method that leverages nearest-neighbor distances, providing a more accurate representation of distances used in OOD detection.
As discussed in the introduction section, distance-based methods struggle to assign low scores to samples located in the regions around ID classes, leading to difficulties in detecting near-OOD samples.
NNGuide \citep{park2023nearest} uses a classifier-based score function to improve KNN's performance on near-OOD samples. However, this approach leads to diminished effectiveness in detecting far-OOD samples.
In contrast, our method not only preserves the good performance of distance-based methods on far-OOD samples, but also enhances their capability to detect near-OOD samples.

\section{Preliminaries}
In this section, we introduce the background of OOD detection task and how the distance-based methods perform the OOD detection.

\subsection{OOD detection task}
OOD detection can be formulated as a binary classification problem. When presented with a novel test input, the model $f$, trained on ID training data, should ascertain whether the input falls within the ID categories or is considered OOD.
Let $\mathbf{x}$ denotes the test input. The OOD detection aims to define a decision function $\mathcal{G}$:
\begin{align}
\begin{split}
\mathcal{G}(\mathbf{x},f)= \left \{
\begin{array}{ll}
    \text{ID} &  \mathcal{S}(\mathbf{x}) \ge \gamma \\
    \text{OOD} & \mathcal{S}(\mathbf{x})<\gamma
\end{array}
\right.
\end{split}
\end{align}
where $\mathcal{S}(\cdot)$ is the score function and $\gamma$ is a chosen threshold to make a large portion of ID data correctly classified. 
Inputs $\mathbf{x}$ with scores $\mathcal{S}(\mathbf{x})$ higher than $\gamma$ are classified as ID, and with lower scores are classified as OOD.
The difficulty of OOD detection is the lack of OOD data, which are unseen in the training process.
This makes it a semi-supervised or unsupervised task.

\subsection{Distance-based OOD detection methods}
\label{distance-based setup}
Distance-based methods calculate OOD scores by measuring the distance between a test sample and the training data. This measurement is typically achieved by using dissimilarity metrics such as Euclidean or Mahalanobis distance. A test sample that is similar to the training data, indicated by a low dissimilarity, will be assigned higher OOD score. Conversely, a sample that exhibits greater dissimilarity from the training data is given lower OOD score, suggesting that it is likely from a different distribution. 

To be more specific, distance-based methods first collect the feature embeddings of the training data to build a training feature bank.
Then during testing, they compare the feature embedding of the test sample against the feature bank to calculate the OOD score.
Let $f(\cdot)$ denote the encoder of a model that maps an input sample $\mathbf{x}$ into a high-dimensional feature vector $\mathbf{z}$. Then the deep feature of $\mathbf{x}$, represented by an $m$-dimensional vector $\mathbf{z} = [z_1,z_2,...,z_m]^\mathsf{T}$, is extracted by $f(\cdot)$.
The deep features of training data are stored in a feature bank $\mathbb{Z}_n$, which can be represented as:
\begin{equation}
    \mathbb{Z}_n = \{ \mathbf{z}_i \}_{i=1}^{n} = \{ f(\mathbf{x}_i) \}_{i=1}^{n}
\end{equation}
where $\mathbf{x}_i$ is the training data and $\mathbf{z}_i =  [z_{i1},z_{i2},...,z_{im}]^\mathsf{T}$ is its feature embedding. The distance between the test sample and the training feature bank is calculated as:
\begin{equation}
    \text{Distance}(  \mathbf{x}) = \text{Dissimilarity}( \, \mathbb{Z}_n,f(\mathbf{x}) \, )
\end{equation}
where $\text{Dissimilarity} ( \cdot , \cdot  )$ is the way to measure the dissimilarity in each distance-based methods.
Based on this distance function, the final OOD score function $\mathcal{S}(\cdot)$ can be defined as:
\begin{equation}
  \mathcal{S}(\mathbf{x}) = h( \,  \text{Distance}(\mathbf{x}) \,  )
\end{equation}
where $h(\cdot)$ is a function that maps the calculated distance to an OOD score. This function correlates low dissimilarity with higher scores, and conversely, high dissimilarity with lower scores.

\section{Method: Feature Bank Enhancement}
Previous distance-based methods primarily focused on designing efficient score functions, but paid less attention to the inherent limitations in the training data.
To address this, we propose \textbf{F}eature \textbf{B}ank \textbf{E}nhancement, a simple yet effective approach that enhances distance-based methods for near-OOD samples while preserving their performance on far-OOD samples.
Based on our observation, some training features are significantly deviated from the center where most training features are clustered. We refer to these clustered features as typical training features, and the deviated ones are termed extreme training features.
Our key idea is to constrain extreme training features to the separation boundaries between these two kinds of training features, thereby mitigating their negative influence on the calculation of OOD scores.

We propose the FBE operation, applied to the training feature bank of distance-based methods.
Consider the setup introduced in Section \ref{distance-based setup}.
We start by calculating the absolute distance vector $ \mathbf{d}_i $ between each training feature vector $\mathbf{z}_i$ and the mean vector $\boldsymbol{\mu} = [\mu_1,\mu_2,...,\mu_m]^\mathsf{T} $ of the feature bank $\mathbb{Z}_n$.
The absolute distance vector $ \mathbf{d}_i $ can be presented as:
\begin{equation}
    \mathbf{d}_i =  [ \, 
                    \lvert  z_{i1} - \mu_1  \rvert, 
                     \lvert  z_{i2} - \mu_2  \rvert, 
                     ...,
                     \lvert  z_{im} - \mu_m  \rvert
                     \, 
                     ]^\mathsf{T} 
\end{equation}
The calculation of $\mathbf{d}_i$ allows us to evaluate the deviation of each feature vector from the center of the feature bank in each dimension. Since the subtraction and absolute value operations are applied to each dimension, $\mathbf{d}_i$  By performing this operation on each training feature vector $\mathbf{z}_i \in \mathbb{Z}_n$, we obtain an absolute distance bank $\mathbb{D}_n = \{ \mathbf{d}_i \}_{i=1}^{n}$.
Then we compute the $\lambda$-th percentile of $\mathbb{D}_n$ in each dimension to establish the boundaries between two kinds of training features $\mathbf{d}^{\ast}$:
\begin{equation}
    \mathbf{d}^{\ast}
    = [d^{\ast}_1,d^{\ast}_2,...,d^{\ast}_m]^\mathsf{T}
    = \text{Percentile}(\mathbb{D}_n , \lambda)
\end{equation}
where $\lambda$ is a tuning parameter, representing the strictness of the deviation boundary.
In this way, we obtain the typical training features regions as $[ \boldsymbol{\mu} - \mathbf{d}^{\ast} , \boldsymbol{\mu} + \mathbf{d}^{\ast} ]$.

\begin{algorithm}[t]
    \fontsize{8pt}{9.5pt}\selectfont
    \caption{Feature Bank Enhancement (FBE)}
    \label{alg1}
    \textbf{Input}: Basic distance-based score function ($\mathcal{S}$), Training feature bank ($\mathbb{Z}_n$)\\
    \textbf{Parameter}: Retention rate of typical training features ($\lambda$)\\
    \textbf{Output}: FBE-improved score function ($\mathcal{S}^{\ast}$)
    
    \begin{algorithmic}[1] 
        \STATE $\boldsymbol{\mu} = \frac{1}{n} \sum_{i=1}^{n} \mathbf{z}_i $ where $ \mathbf{z}_i \in \mathbb{Z}_n$
        \STATE $ \mathbb{D} =  \{ \mathbf{d}_i  \}_{i=1}^{n}
                 = \{ [ \, 
                    \lvert  z_{i1} - \mu_1  \rvert, 
                     ...,
                     \lvert  z_{im} - \mu_m  \rvert
                     \, 
                     ]^\mathsf{T}  \}_{i=1}^{n}      $
        \STATE $\mathbf{d}^{\ast} = \text{Percentile}(\mathbb{D}_n , \lambda)$
\FOR{each $\mathbf{z}_i \in \mathbb{Z}_n$}
    \FOR{each dimension $j$ of $\mathbf{z}_i$}
        \IF{${z}_{ij} > \mu_j + d^{\ast}_j$} 
            \STATE ${z}_{ij}^{\ast} = \mu_j + d^{\ast}_j$
        \ELSIF{${z}_{ij} < \mu_j - d^{\ast}_j$}
            \STATE ${z}_{ij}^{\ast} = \mu_j - d^{\ast}_j$
        \ELSE
            \STATE ${z}_{ij}^{\ast} = {z}_{ij}$
        \ENDIF
    \ENDFOR
    \STATE $\mathbf{z}_i^{\ast} =[{z}_{i1}^{\ast},...,{z}_{im}^{\ast}]^\mathsf{T}$
\ENDFOR
\STATE $\mathbb{Z}_n^{\ast} = \{   \mathbf{z}_i^{\ast}   \}_{i=1}^{n}  $
\STATE $\mathcal{S}^{\ast}(\mathbf{x}) = \mathcal{S}(\mathbf{x})|_{\mathbb{Z}_n = \mathbb{Z}_n^{\ast}}$
\STATE \textbf{return} $\mathcal{S}^{\ast}$
\end{algorithmic}
\end{algorithm}

Next, we examine the original feature vector $\mathbf{z}_i$ to determine whether it corresponds to typical or extreme training feature in each dimension, and apply a transformation that constrains extreme training features to the boundaries:
\begin{equation}
    {z}_{ij}^{\ast} = 
\left \{
\begin{array}{ll}
    \mu_j + d^{\ast}_j          & {z}_{ij} > \mu_j + d^{\ast}_j  \\
    \quad {z}_{ij}              & {z}_{ij}\in[ \, \mu_j - d^{\ast}_j , \mu_j + d^{\ast}_j \,  ] \\
    \mu_j - d^{\ast}_j          & {z}_{ij} < \mu_j - d^{\ast}_j
\end{array}
\right.
\end{equation}
This transformation results in new feature vectors $\mathbf{z}_i^{\ast} =[{z}_{i1}^{\ast},{z}_{i2}^{\ast},...,{z}_{im}^{\ast}]^\mathsf{T}$.
Following this procedure, a refined feature bank $\mathbb{Z}_n^{\ast}$ is generated.
By combining this refined feature bank $\mathcal{S}^{\ast}$ with basic distance-based methods, the FBE-improved score function $\mathcal{S}^{\ast}(\cdot)$ can be represented as:

\begin{equation}
    \mathcal{S}^{\ast}(\mathbf{x}) = h( \,  \text{Dissimilarity}( \,  \mathbb{Z}_n^{\ast},f(\mathbf{x}) \,  ) \,  )
\end{equation}
We further summarize our framework in Algorithm \ref{alg1}.

\section{Experiments}
In this section, we first introduce the experimental setup in Section \ref{setup}.
Then we verify the effectiveness of our method in enhancing the fine-grained detection capability of distance-based methods in Section \ref{effectiveness}, and proceed to present the primary experimental results in Section \ref{main results}.
Finally, we evaluate the compatibility of FBE with network truncation method in Section \ref{exp:compatibility}.


\subsection{Setup}
\label{setup}

\textbf{Benchmarks.} We choose the widely used large-scale ImageNet-1k \citep{deng2009imagenet} and small-scale CIFAR-10 \citep{krizhevsky2009learning} as the benchmarks. The \textbf{Large-scale ImageNet-1k benchmark} provides a challenging and realistic application scenario, making the experimental results more representative and relevant for real-world applications. 
We use ImageNet-1k \citep{deng2009imagenet} as the ID dataset, and OOD datasets include subsets from iNaturalist \citep{van2018inaturalist}, Places \citep{zhou2017places}, SUN \citep{xiao2010sun}, Textures \citep{cimpoi2014describing}, and OpenImage-O \citep{wang2022vim}, with no category overlaps.
As defined by OpenOOD \citep{yang2022openood}, iNaturalist represents a near-OOD dataset and Textures represents a far-OOD dataset. Following previous works, we use ResNet-50 \citep{he2016deep} as the backbone model. The model is trained on the training fold of ImageNet-1k, and the classification layers are prohibited to see any instance from OOD datasets.
For the \textbf{CIFAR-10 benchmark}, CIFAR-10 is the ID dataset, with OOD datasets including SVHN \citep{netzer2011reading}, LSUN \citep{yu2015lsun}, iSUN \citep{xu2015turkergaze}, Textures \citep{cimpoi2014describing}, and Places365 \citep{zhou2017places}.
Consistent with previous settings, we use ResNet-18 \citep{he2016deep} as the backbone model.

\textbf{Baselines.}
We compare our method with a range of state-of-the-art OOD detection methods, including \textsl{distance-based methods}: SSD \citep{sehwag2021ssd}, ViM \citep{wang2022vim}, KNN \citep{sun2022out} and NNGuide \citep{park2023nearest}; \textsl{classifier-based methods}: MSP \citep{hendrycks2016baseline}, Energy \citep{liu2020energy}, DML+ \citep{zhang2023decoupling}, MaxLogit and KL \citep{hendrycks2019scaling}; and \textsl{network truncation methods}: ODIN \citep{liang2017enhancing}, GODIN \citep{hsu2020generalized}, DICE \citep{sun2022dice}, ReAct \citep{sun2021react}, RankFeat \citep{song2022rankfeat}, BATS \citep{zhu2022boosting}, LAPS \citep{he2024exploring}, and ASH \citep{djurisic2022extremely}.

\textbf{Metrics.}
We evaluate performance using the FPR95 and AUROC metrics.
FPR95 measures the false positive rate at a 95\% true positive rate. AUROC represents the area under the receiver operating characteristic curve, indicating the probability that an ID sample will receive a higher score than an OOD sample. Lower FPR95 and higher AUROC values correspond to better performance.

\subsection{Improvements on distance-based methods}
\label{effectiveness}

\begin{figure}[t]
    \centering

    \begin{subfigure}[b]{0.156\textwidth}
        \centering
        \includegraphics[width=\textwidth]{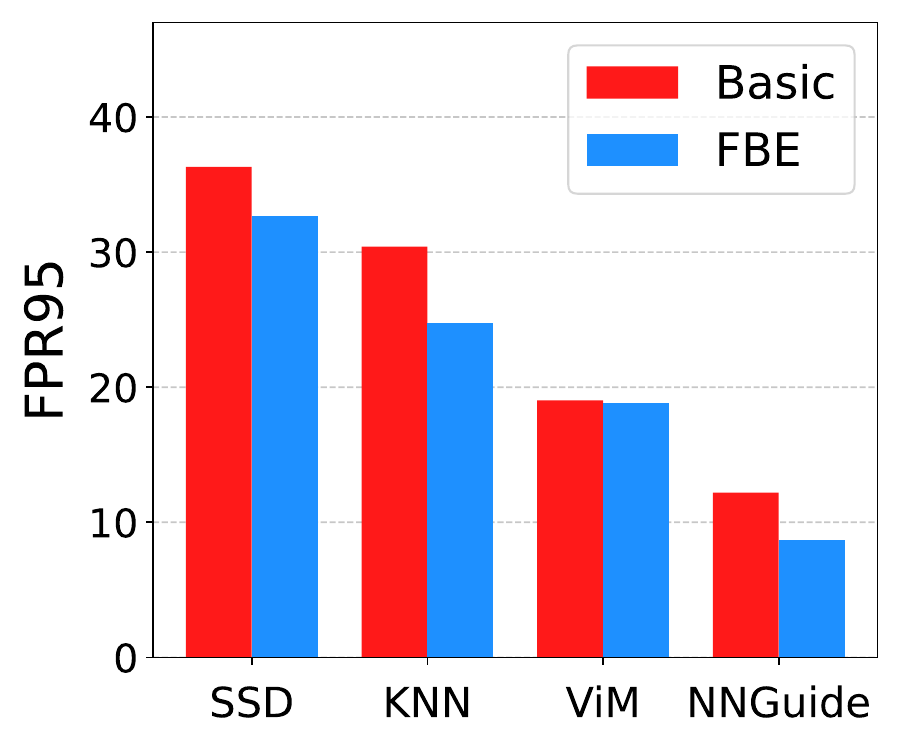}
        \caption{Near-OOD dataset}
        \label{fig3a}
    \end{subfigure}
    \begin{subfigure}[b]{0.156\textwidth}
        \centering
        \includegraphics[width=\textwidth]{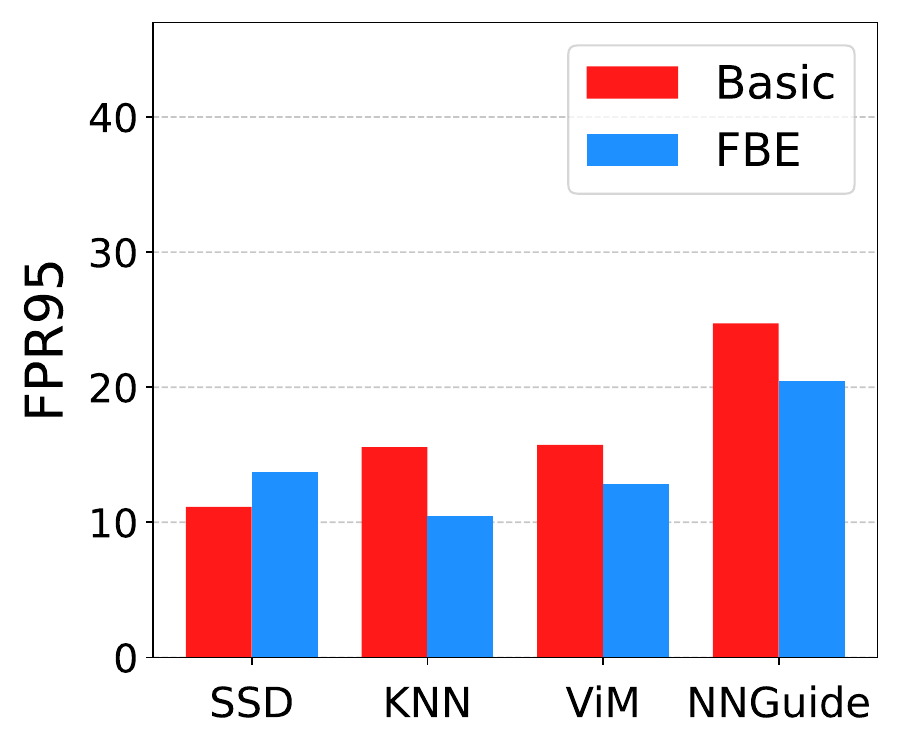}
        \caption{Far-OOD dataset}
        \label{fig3b}
    \end{subfigure}
    \begin{subfigure}[b]{0.156 \textwidth}
        \centering
        \includegraphics[width=\textwidth]{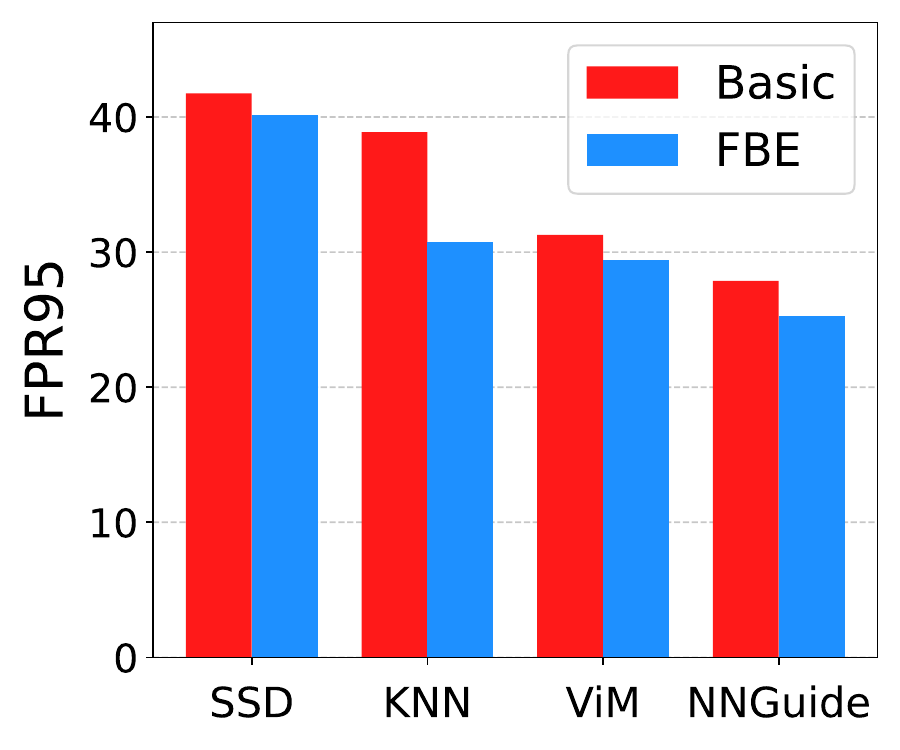}
        \caption{Average performance}
        \label{fig3c}
    \end{subfigure}

    \caption{\textbf{Apply FBE on distance-based methods.} The performance of various distance-based methods on near-OOD dataset, far-OOD dataset and the average performance in ImageNet-1k benchmark.
    Red rectangles denote basic distance-based methods, and blue rectangles denote their FBE-improved counterparts.}
    \label{fig3}

\end{figure}

\begin{table*}[t]
    \centering
    \fontsize{7.5pt}{9pt}\selectfont
    \resizebox{\textwidth}{!}{
    \setlength{\tabcolsep}{0.5 mm} 
    \begin{tabular}{p{19mm}|cc|cc|cc|cc|cc|cc}
    \toprule[0.5mm]
    
    \raisebox{0mm}[0mm][1.5mm]{\multirow{3}{*}{Method}} & \multicolumn{12}{c}{OOD Datasets} \\
    
    & \multicolumn{2}{c|}{iNaturalist \citep{van2018inaturalist}} 
    & \multicolumn{2}{c|}{SUN \citep{xiao2010sun}} 
    & \multicolumn{2}{c|}{Places \citep{zhou2017places}} 
    & \multicolumn{2}{c|}{Textures \citep{cimpoi2014describing}}
    & \multicolumn{2}{c|}{OpenImage-O \citep{wang2022vim}}
    & \multicolumn{2}{c}{Average}\\
    
    & {FPR95} & {AUROC} 
    & {FPR95} & {AUROC} 
    & {FPR95} & {AUROC}  
    & {FPR95} & {AUROC}  
    & {FPR95} & {AUROC}  
    & {FPR95} & {AUROC}\\

    \midrule[0.2mm]

    MSP \citep{hendrycks2016baseline}            
                    & 29.70 & 93.78
                    & 59.66 & 84.54
                    & 61.01 & 84.25
                    & 49.88 & 84.90
                    & 48.26 & 89.37 
                    & 49.70 & 87.37 \\
    
    Energy \citep{liu2020energy}          
                    & 20.94 & 96.17
                    & 46.90 & 88.89
                    & 51.15 & 87.68
                    & 39.38 & 88.90
                    & 42.17 & 92.11
                    & 40.11 & 90.75 \\
    
    DML+ \citep{zhang2023decoupling}            
                    & 21.44 & 96.06
                    & 48.69 & 88.73
                    & 52.04 & 87.64
                    & 43.19 & 88.14
                    & 42.70 & 91.90
                    & 41.61 & 90.49 \\
    
    MaxLogit \citep{hendrycks2019scaling}        
                    & 22.01 & 95.99
                    & 50.92 & 88.41
                    & 53.88 & 87.35
                    & 42.25 & 88.43
                    & 42.75 & 92.01
                    & 42.36 & 90.44 \\
    
    KL \citep{hendrycks2019scaling}             
                    & 20.94 & 96.17
                    & 46.89 & 88.89
                    & 51.14 & 87.68
                    & 39.36 & 88.90
                    & 42.17 & 92.11
                    & 40.10 & 90.75 \\
    
    SSD \citep{sehwag2021ssd}            
                    & 36.31 & 94.34
                    & 56.35 & 88.28
                    & 65.45 & 84.46
                    & \textbf{11.15} & \textbf{97.00}
                    & 39.55 & 92.81
                    & 41.76 & 91.38 \\
    
    ViM \citep{wang2022vim}             
                    & 19.02 & 96.57
                    & 40.47 & 90.65
                    & 50.13 & 88.27
                    & 15.73 & 94.37
                    & 31.06 & \textbf{94.16}
                    & 31.28 & 92.80 \\
    
    KNN \citep{sun2022out}             
                    & 30.40 & 94.71
                    & 51.12 & 88.06 
                    & 60.71 & 84.45
                    & 15.57 & 95.42
                    & 36.61 & 92.80
                    & 38.88 & 91.09 \\
    
    NNGuide \citep{park2023nearest}        
                    & 12.19 & 97.38 
                    & 31.20 & 91.71
                    & 38.15 & 90.22
                    & 24.73 & 91.72
                    & 33.11 & 93.23
                    & 27.88 & 92.85 \\
    
    \rowcolor[gray]{0.8} FBE (ours)
                    & \textbf{8.72} & \textbf{98.12}
                    & \textbf{29.64} & \textbf{92.46}
                    & \textbf{37.78} & \textbf{90.71}
                    & 20.43 & 93.26
                    & \textbf{29.88} & 94.09
                    & \textbf{25.29} & \textbf{93.73} \\      
    \bottomrule[0.5mm]
    \end{tabular}}
    \vspace{3pt} 
    \caption{\textbf{Main Results on ImageNet-1k benchmark.}
    \label{table:main results on imagenet} Higher AUROC values and lower FPR95 values indicate better performance. The best results are emphasized in \textbf{bold}.}
\end{table*}

\begin{table*}[t]
    \centering
    \fontsize{7.5pt}{9pt}\selectfont
    \resizebox{\textwidth}{!}{
    \setlength{\tabcolsep}{0.5 mm} 
    \begin{tabular}{p{19mm}|cc|cc|cc|cc|cc|cc}
    \toprule[0.5mm]
    
    \raisebox{0mm}[0mm][1.5mm]{\multirow{3}{*}{Method}} & \multicolumn{12}{c}{OOD Datasets} \\
    
    & \multicolumn{2}{c|}{SVHN \citep{netzer2011reading}} 
    & \multicolumn{2}{c|}{LSUN \citep{yu2015lsun}} 
    & \multicolumn{2}{c|}{iSUN \citep{xu2015turkergaze}} 
    & \multicolumn{2}{c|}{Texture \citep{cimpoi2014describing}}
    & \multicolumn{2}{c|}{Places365 \citep{zhou2017places}}
    & \multicolumn{2}{c}{Average}\\
    
    & {FPR95} & {AUROC} 
    & {FPR95} & {AUROC} 
    & {FPR95} & {AUROC}  
    & {FPR95} & {AUROC}  
    & {FPR95} & {AUROC}  
    & {FPR95} & {AUROC}\\

    \midrule[0.2mm]

    MSP \citep{hendrycks2016baseline}            & 7.68 & 98.60 
                                                & 3.69 & 99.15 
                    & 17.58 & 97.25 
                    & 18.09 & 97.09 
                    & 27.69 & 95.00 
                    & 14.95 & 97.42 \\
    
    Energy \citep{liu2020energy}          & 4.04 & 99.11 
                    & 1.39 & 99.55 
                    & 9.65 & 98.10 
                    & 13.35 & 97.63 
                    & 21.32 & 95.79 
                    & 9.95 & 98.04 \\
    
    DML+ \citep{zhang2023decoupling}            & 4.96 & 98.96 
                    & 1.51 & 99.51 
                    & 9.88 & 98.07
                    & 86.74 & 97.03 
                    & 23.29 & 95.56 
                    & 11.35 & 97.83 \\
    
    MaxLogit \citep{hendrycks2019scaling}        & 5.03 & 98.94 
                    & 1.92 & 99.39 
                    & 11.56 & 97.85 
                    & 14.75 & 97.48 
                    & 23.06 & 95.59
                    & 11.26 & 97.85 \\
    
    KL \citep{hendrycks2019scaling}              & 4.06 & 99.10 
                    & 1.39 & 99.54
                    & \textbf{9.51} & \textbf{98.12} 
                    & 13.30 & 97.64 
                    & 21.19 & 95.80 
                    & 9.89 & 98.04 \\
    
    SSD \citep{sehwag2021ssd}             & \textbf{0.18} & \textbf{99.93} 
                    & 1.30 & 99.52 
                    & 37.54 & 94.64 
                    & \textbf{5.41} & \textbf{98.94} 
                    & 21.81 & 95.49 
                    & 13.25 & 97.70 \\
    
    ViM \citep{wang2022vim}             & 3.07 & 99.42 
                    & 46.69 & 93.56 
                    & 83.24 & 86.74 
                    & 39.84 & 94.41 
                    & 55.30 & 90.89 
                    & 45.63 & 93.00 \\
    
    KNN \citep{sun2022out}             & 1.23 & 99.70 
                    & 1.78 & 99.48
                    & 20.00 & 96.72 
                    & 8.14 & 98.56 
                    & 22.23 & 95.56 
                    & 10.68 & 98.00 \\
    
    NNGuide \citep{park2023nearest}        & 3.27 & 99.34 
                    & 1.24 & 99.60
                    & 11.57 & 97.87 
                    & 10.74 & 98.18 
                    & 21.33 & 95.78 
                    & 9.63 & 98.15 \\
    
    \rowcolor[gray]{0.8} FBE (ours)
                    & 2.14 & 99.55
                    & \textbf{0.81} & \textbf{99.74}
                    & 10.25 & 97.95
                    & 10.04 & 98.27
                    & \textbf{20.34} & \textbf{95.86}
                    & \textbf{8.72} & \textbf{98.27} \\      
    \bottomrule[0.5mm]
    \end{tabular}}
    \vspace{3pt} 
    \caption{\textbf{Main Results on CIFAR-10 benchmark.}\label{table:main results on cifar} Higher AUROC values and lower FPR95 values indicate better performance. The best results are emphasized in \textbf{bold}.}
\end{table*}

We first evaluate the effectiveness of FBE on many distance-based methods.
In Figure \ref{fig3}, we compare each basic distance-based method with its FBE-modified version in ImageNet-1k benchmark.
On the near-OOD dataset (Figure \ref{fig3a}), which typically presents challenges for distance-based methods, FBE significantly improves their performance.
For example, KNN's FPR95 decreases by 5.65\%, and NNGuide achieves an even lower FPR95.
On the far-OOD dataset (Figure \ref{fig3b}), where distance-based methods already perform well, our FBE either maintains or slightly improves their effectiveness.
Comprehensive evaluations across all datasets (Figure \ref{fig3c}) demonstrate an overall improvement in the detection capabilities of distance-based methods, with KNN showing the most notable improvement. NNGuide and ViM, which already exhibit strong performance, also benefit from further enhancement through FBE.
These enhancements across various distance-based methods demonstrate the effectiveness of FBE. We select NNGuide, the most effective among these methods, as the basic distance-based method to compare our FBE with other OOD score functions in the subsequent section.

\subsection{Main Results}
\label{main results}

In Table \ref{table:main results on imagenet}, we compare our method with extensive collection of competitive score function detection methods on the large-scale ImageNet-1k benchmark.
Our method achieves a new overall \textbf{state-of-the-art} performance, outperforming the second-best baseline by 2.59\% in FPR95 and 0.88\% in AUROC.
More specifically, FBE achieves the best performance on the iNaturalist, SUN, and Places OOD datasets across both metric, and on OpenImage-O under the FPR95 metric.
On the near-OOD dataset (iNaturalist), FBE outperforms the second-best baseline by 3.47\% in FPR95 and 0.79\% in AUROC, which demonstrates our method can properly improve the fine-grained detection capabilities on near-OOD samples.

We further evaluate our method on the small-scale CIFAR-10 benchmark in Table \ref{table:main results on cifar}.
Although many methods already perform well on this smaller benchmark, Table \ref{table:main results on cifar} shows that our method also outperforms all baselines, leading by 0.91\% in FPR95 and 0.12\% in AUROC on average performance. 

These results showcase the versatility and robustness of FBE across different contexts, demonstrating that our method achieves superior OOD detection performance on almost all experimental settings.

\subsection{Comparison with network truncation methods}
\label{exp:compatibility}

\begin{table}[t]
    \centering
    \fontsize{7.5pt}{9pt}\selectfont
    \setlength{\tabcolsep}{2mm} 
    \begin{tabular}{llcc}
        \toprule[0.5mm]
        Detection Method & Backbone Model & FPR95 & AUROC \\
        \midrule
        ODIN \citep{liang2017enhancing} *         & ResNet-50      & 56.48 & 85.41 \\
        GODIN \citep{hsu2020generalized} *        & ResNet-50      & 66.07 & 82.02 \\
        DICE \citep{sun2022dice} *                & ResNet-50      & 34.75 & 90.77 \\
        ReAct \citep{sun2021react} + DICE \citep{sun2022dice} *  & ResNet-50  & 27.25 & 93.40 \\
        RankFeat \citep{song2022rankfeat} *       & ResNet-101     & 36.80 & 92.15 \\
        BATS \citep{zhu2022boosting} *            & ResNet-50      & 27.11 & 94.28 \\
        LAPS \citep{he2024exploring} *            & ResNet-50      & 23.68 & 94.78 \\
        ASH \citep{djurisic2022extremely} *       & ResNet-50      & 22.73 & 95.06 \\
        ReAct (+ Energy) \citep{sun2021react} *                 & ResNet-50  & 31.43 & 92.95 \\
        ReAct \citep{sun2021react} + MSP \citep{hendrycks2016baseline}          
        & ResNet-50      & 56.08 & 87.24 \\
        ReAct \citep{sun2021react} + DML+ \citep{zhang2023decoupling}            
        & ResNet-50      & 30.68 & 92.99 \\
        ReAct \citep{sun2021react} + MaxLogit \citep{hendrycks2019scaling}        
        & ResNet-50      & 40.13 & 91.80 \\
        ReAct \citep{sun2021react} + KL \citep{hendrycks2019scaling}              
        & ResNet-50      & 32.77 & 93.07 \\
        ReAct \citep{sun2021react} + SSD \citep{sehwag2021ssd}             
        & ResNet-50      & 55.68 & 83.97 \\
        ReAct \citep{sun2021react} + ViM \citep{wang2022vim}             
        & ResNet-50      & 25.27 & 94.97 \\
        ReAct \citep{sun2021react} + KNN \citep{sun2022out}             
        & ResNet-50      & 39.11 & 90.69 \\
        ReAct \citep{sun2021react} + NNGuide \citep{park2023nearest}         
        & ResNet-50      & 18.85 & 95.65 \\
        \rowcolor[gray]{0.8} ReAct \citep{sun2021react} + FBE (ours)      & ResNet-50 & \textbf{17.89} & \textbf{95.83} \\ 
        \bottomrule[0.5mm]
    \end{tabular}
    \vspace{3pt} 
    \caption{\textbf{Comparison with network truncation methods.} We reported the average performance across iNaturalist, SUN, Places, and Textures, four OOD datasets. * indicates that the results are taken from the original references.}
    \label{table:network truncation}
\end{table}

Besides score function methods, recent works show that network truncation methods can also achieve excellent performance in OOD detection tasks. These approaches focus on modifying certain network layers to perturb output signals, significantly affecting OOD sample signals while retaining those of ID samples. However, such methods cannot be used alone and need to be combined with score functions to realize OOD detection capabilities. 

We demonstrate the compatibility of our method with network truncation methods by combining FBE with ReAct, and comparing with other recently proposed network truncation methods.
Table \ref{table:network truncation} shows that when combined with ReAct, our proposed FBE achieves the best performance across both FPR95 and AUROC metrics, outperforming other recently proposed network truncation methods.
Moreover, when comparing ReAct combined with various score functions, FBE also demonstrates greater effectiveness and relevance than the other methods.

\section{Analysis and Discussion}
In this section, we delve deeper into the working mechanisms through theoretical analysis and supplementary experiments.
We begin with a theoretical examination that constraining extreme training features may increase the probability of ID sample receiving higher scores than OOD samples.
Subsequently, we explore the influence of the hyperparameter $\lambda$ in FBE.
Then we discuss the efficiency of FBE. 
In the last part, we point out the limitation of our FBE approach.

\subsection{Theoretical Analysis}
Here we aim to demonstrate that the constraint operation in FBE enhances the probability of ID samples receiving higher scores than OOD samples.
Following the mathematical framework in \citep{sun2021react}, let $\mathcal{S}(\cdot)$ and $\mathcal{S}^{\ast}(\cdot)$ denote the original and FBE-improved distance-based score functions, respectively.
Therefore, we aim to demonstrate:
\begin{equation}
    \Pr\{ \; \mathcal{S}(x_{\text{in}}) > \mathcal{S}(x_{\text{out}}) \; \}  <  \Pr\{ \; \mathcal{S}^{\ast}(x_{\text{in}}) > \mathcal{S}^{\ast}(x_{\text{out}}) \; \}
    \label{equ:enhance1}
\end{equation}
Given that distance-based methods primarily derive their OOD scores by calculating the dissimilarity between the high-dimensional features of input samples and the training features, with lower dissimilarity leading to higher scores, we can express this relationship more concretely. If Euclidean distance is used to calculate the dissimilarity, the expression in Formula \ref{equ:enhance1} can be transformed to:

\begin{equation}
    \Pr\{ \; d_{\text{in}} < d_{\text{out}} \; \}  <  \Pr\{ \; d_{\text{in}}^{\ast} < d_{\text{out}}^{\ast} \; \}
    \label{equ:two prob}
\end{equation}
where $d_{\text{in/out}} = \Vert \mathbf{z_{\text{train}} - z_{\text{in/out}}} \Vert_2$ is the Euclidean distance between the test feature and the training feature, and $d_{\text{in/out}}^{\ast}$ is its FBE-improved counterpart.

Following \citep{sun2021react}, we use Gaussian and epsilon-skew-normal distributions to model the distributions of ID features and OOD features respectively.
More specifically, considering a feature vector $\mathbf{z} = [z_1,z_2,...,z_m]^\mathsf{T}$, each ID feature $z_{\text{in}}$ follows a Gaussian distribution $\mathcal{N}(\mu, \sigma_{\text{in}}^2)$, with $\sigma_{\text{in}} > 0$.
Regarding OOD features, Sun \emph{et al}. observed that the mean value $\mu$ exhibits strong positive skewness (\emph{i.e.,} the right tail is much denser than the left tail), which is consistent across different datasets and model architectures \citep{sun2021react}.
Based on this empirical finding, they model the OOD feature $z_{\text{out}}$ by the epsilon-skew-normal (ESN) distribution \citep{mudholkar2000epsilon}.
It is assumed that $z_{\text{out}} \sim \text{ESN}(\mu, \sigma_{\text{out}}^2, \epsilon)$, with $\epsilon \in [-1,1]$ controlling the skewness.
The ESN distribution is positively-skewed when $\epsilon < 0$.
Its density function is given by:
\begin{equation}
    q({z_{\text{out}}}) = 
\left \{
\begin{array}{ll}
\phi((z_{\text{out}} - \mu) / \sigma_{\text{out}} (1 + \epsilon)) / \sigma_{\text{out}}   &  z_{\text{out}} < \mu  \\
\phi((z_{\text{out}} - \mu) / \sigma_{\text{out}} (1 + \epsilon)) / \sigma_{\text{out}}   &  z_{\text{out}} \ge \mu
\end{array}
\right.
\end{equation}
where $\phi(\cdot)$ is the density function of standard Gaussian distribution.
It should be emphasized that $\sigma_{\text{out}}$ should be larger than $\sigma_{\text{in}}$, corresponding to the assumption that OOD features are notably separated from ID features.

After applying FBE, the distribution of the training feature is limited.
The rectified training feature $z_{\text{train}}^{\ast}$ follows a modified normal distribution $\mathcal{N}^{\ast}(\mu, \sigma_{\text{in}}^2)$, which is defined as $\mathcal{N}^{L} = \text{max}( \text{min}(\mathcal{N}(\mu, \sigma_{\text{in}}^2) , \mu + \lambda) ,   \mu - \lambda  )$.
Then the modified distances $d_{\text{in}}^{\ast}$ and $d_{\text{out}}^{\ast}$ can be presented as:
\begin{equation}
    d_{\text{in/out}}^{\ast} = \Vert \mathbf{z_{\text{train}}^{\ast}-z_{\text{in/out}}} \Vert_2 = \sqrt{\sum_{i=1}^{m}(z_{\text{train},i}^{\ast} - z_{\text{in/out},i})^2}
    \label{equ:distance}
\end{equation}

\begin{figure}[t]
    \centering
    \includegraphics[width=0.35\textwidth]{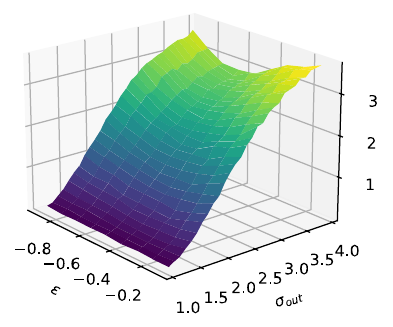}
    \vspace{15pt} 
    \caption{\textbf{Monte Carlo Simulation Results.} The percentage value of $\Pr\{ \; d_{\text{in}}^{\ast} < d_{\text{out}}^{\ast} \; \}  -  \Pr\{ \; d_{\text{in}} < d_{\text{out}} \; \}$ across skewness parameter $\epsilon$ and chaotic-ness parameter $\sigma_{\text{out}}$.
    The positive value correspond to $\Pr\{ \; d_{\text{in}}^{\ast} < d_{\text{out}}^{\ast} \; \}  <  \Pr\{ \; d_{\text{in}} < d_{\text{out}} \; \}$, which suggests that FBE can improve the probability of ID samples receiving higher scores than OOD scores.
    }
    \label{fig4}
\end{figure}

By substituting the expressions of $d_{\text{in}}, d_{\text{out}}, d_{\text{in}}^{\ast}, d_{\text{out}}^{\ast}$ into Formula \ref{equ:two prob}, we can obtain the specific calculation formulas for $\Pr\{ \; d_{\text{in}} < d_{\text{out}} \; \}$ and $\Pr\{ \; d_{\text{in}}^{\ast} < d_{\text{out}}^{\ast} \; \}$.
Using Monte Carlo simulations \citep{metropolis1949monte}, we compute the specific value of these two probabilities.
Figure \ref{fig4} shows a plot of the difference $\Pr\{ \; d_{\text{in}}^{\ast} < d_{\text{out}}^{\ast} \; \}  -  \Pr\{ \; d_{\text{in}} < d_{\text{out}} \; \}$ across varying $\sigma_{\text{out}}$ and $\epsilon$ values.
We set $\sigma_{\text{in}}$ to 1 and $\lambda$ to 1.96, indicating that 5\% of the training features, which are far away from the mean value, are identified as extreme features and are constrained.
Observe that the entire curved surface is observed to be above the $z=0$ plane, illustrating the Formula \ref{equ:two prob} is valid.
This result suggests that our FBE can indeed enhance the probability of ID samples receiving higher scores than OOD samples for distance-based methods.

\subsection{Analysis of hyperparameter}
\begin{figure}[t]
    \centering
    \captionsetup{skip=0pt}
    \begin{subfigure}[b]{0.22\textwidth}
        \centering
        \includegraphics[width=\textwidth]{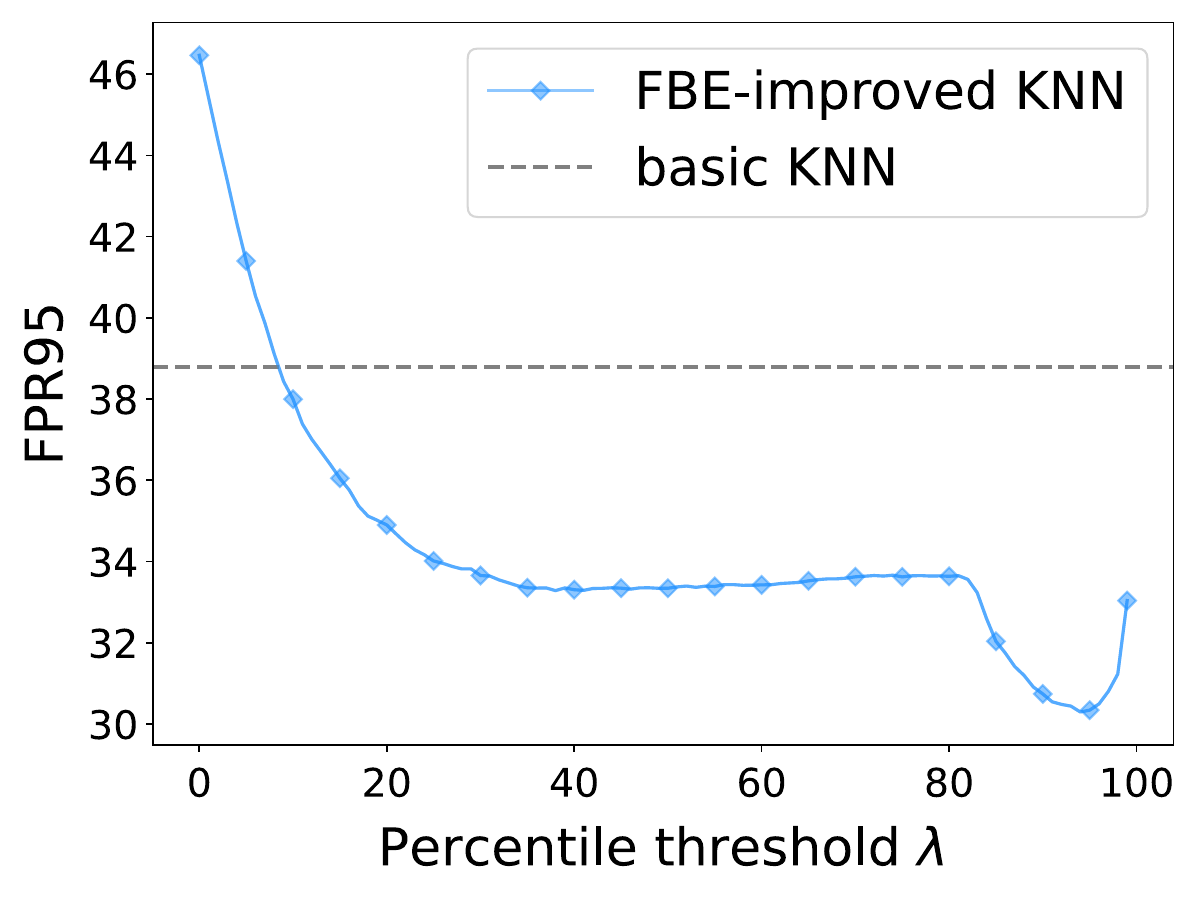}
        \label{fig5a}
    \end{subfigure}
    \hspace{0.01\textwidth}
    \begin{subfigure}[b]{0.22\textwidth}
        \centering
        \includegraphics[width=\textwidth]{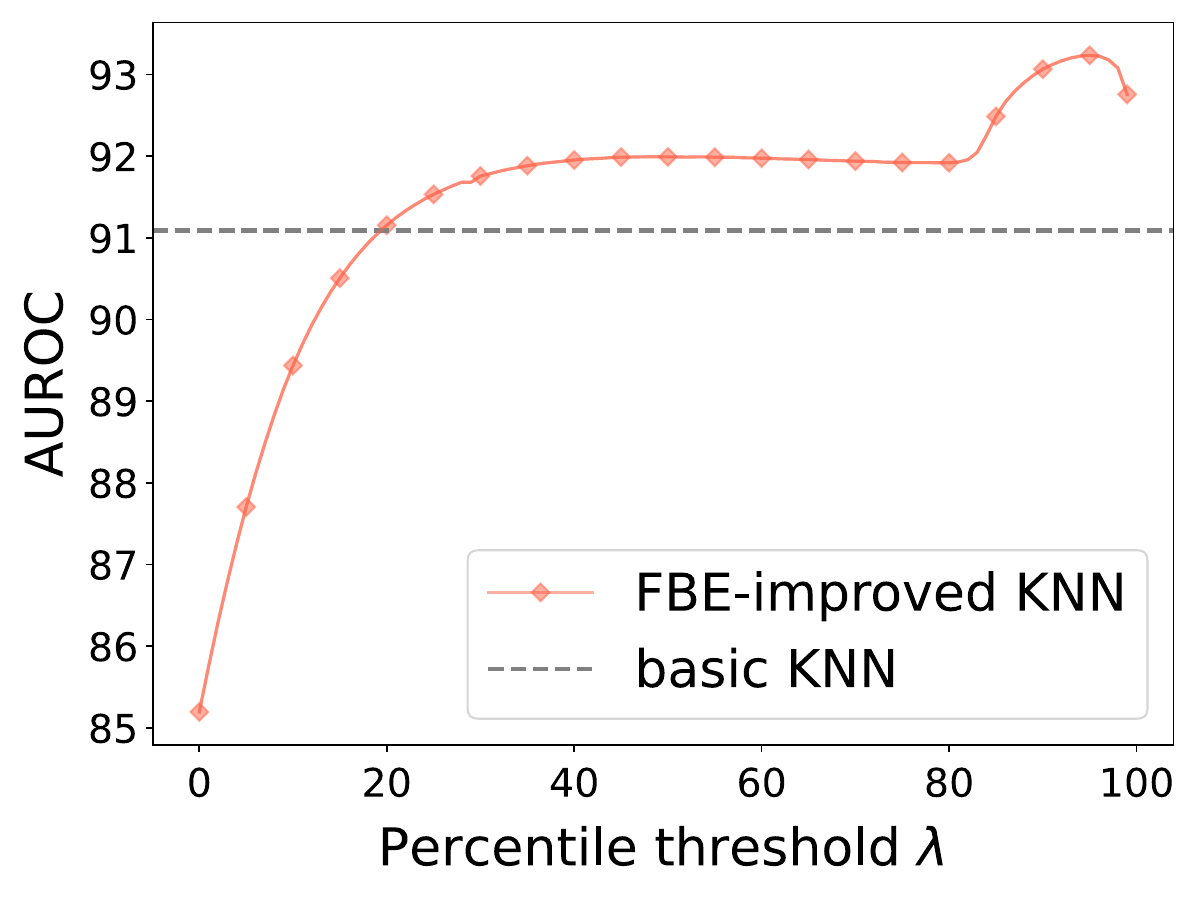}
        \label{fig5b}
    \end{subfigure}
    \caption{\textbf{Influence of the hyperparameter $\lambda$.} 
    \label{fig5}
    \label{analysis:lambda}}
\end{figure}

The hyperparameter $\lambda$, ranging from 0 to 100, controls the strictness of constraints on the training features. We select KNN \citep{sun2022out} as the basic distance-based method due to its classical distance-based properties.
The influence of $\lambda$ on the detection performance is empirically illustrated in Figure \ref{analysis:lambda}.
As $\lambda$ increases, the performance first improves, then declines as $\lambda$ approaches to 100.
On one side, a higher $\lambda$ allows some extreme training features to remain unconstrained, leading to suboptimal results.
On the other side, a lower $\lambda$ leads to more features being identified as extreme training features and thus constrained, which may lead to a loss of valuable information.
Therefore, selecting an optimal $\lambda$ involves balancing the limitation of extreme training features and the preservation of effective information.



\subsection{Efficiency}

\begin{table}[t]
    \centering
    \fontsize{7pt}{9pt}\selectfont
    \setlength{\tabcolsep}{3 mm}{\begin{tabular}{l|ccc}
    \toprule[0.3mm]
    Method & Basic & FBE-improved & Increased time \\ 
    \midrule[0.1mm]
    SSD                     & 74.71 s   & 79.85 s   & +5.14 s \\
    ViM                     & 53.99 s   & 59.58 s   & +5.59 s \\
    KNN                     & 258.18 s  & 262.64 s  & +4.46 s \\ 
    NNGuide                 & 258.65 s  & 261.67 s  & +3.02 s \\
    \bottomrule[0.3mm]
	\end{tabular}}
    \hspace{0.1mm}
    \caption{\textbf{Total processing time.}}
    \label{analysis:efficacy}
\end{table}

Table \ref{analysis:efficacy} shows the total processing time of four basic distance-based methods and their FBE-improved version across five OOD datasets in the ImageNet-1k benchmark. 
Since our approach builds upon distance-based methods, and the FBE operation only involves adjustments to the training feature bank, the efficiency of FBE largely depends on the basic distance-based method. 
For all distance-based methods, the FBE operation adds only few seconds to the total processing time. Therefore, the FBE operation does not significantly affect the detection efficiency of distance-based methods.

\subsection{Limitations}  
The effectiveness of FBE is somewhat constrained when applying to the methods which posit strong assumptions, such as SSD \citep{sehwag2021ssd}.
As depicted in Figure \ref{fig3}, SSD exhibits significantly less improvement compared to other methods.
This might be attributed to the mismatch between the boundaries between typical and extreme training features employed in FBE, and the mismatch of inherent assumptions of these methods.

\section{Conclusion}
In this paper, we propose Feature Bank Enhancement, a simple yet effective method to improve distance-based OOD detection methods by identifying and constraining extreme training features within the training feature bank, bringing them closer to the center where most typical training features cluster. 
By mitigating the negative influence caused by these extreme training features, FBE improves the detection capabilities of distance-based methods.
We not only evaluate the effectiveness of FBE in enhancing various distance-based methods, but also demonstrates its superior performance through extensive experiments on both large-scale ImageNet-1k and small-scale CIFAR-10 OOD benchmarks.
Moreover, we conduct theoretical analysis and supplementary experiments are to provide deeper insights into our approach.

\bibliography{main}

\begin{thebibliography}{38}
\providecommand{\natexlab}[1]{#1}
\providecommand{\url}[1]{\texttt{#1}}
\expandafter\ifx\csname urlstyle\endcsname\relax
  \providecommand{\doi}[1]{doi: #1}\else
  \providecommand{\doi}{doi: \begingroup \urlstyle{rm}\Url}\fi

\bibitem[Cimpoi et~al.(2014)Cimpoi, Maji, Kokkinos, Mohamed, and Vedaldi]{cimpoi2014describing}
M.~Cimpoi, S.~Maji, I.~Kokkinos, S.~Mohamed, and A.~Vedaldi.
\newblock Describing textures in the wild.
\newblock In \emph{Proceedings of the IEEE conference on computer vision and pattern recognition}, pages 3606--3613, 2014.

\bibitem[Deng et~al.(2009)Deng, Dong, Socher, Li, Li, and Fei-Fei]{deng2009imagenet}
J.~Deng, W.~Dong, R.~Socher, L.-J. Li, K.~Li, and L.~Fei-Fei.
\newblock Imagenet: A large-scale hierarchical image database.
\newblock In \emph{2009 IEEE conference on computer vision and pattern recognition}, pages 248--255. Ieee, 2009.

\bibitem[Djurisic et~al.(2022)Djurisic, Bozanic, Ashok, and Liu]{djurisic2022extremely}
A.~Djurisic, N.~Bozanic, A.~Ashok, and R.~Liu.
\newblock Extremely simple activation shaping for out-of-distribution detection.
\newblock \emph{arXiv preprint arXiv:2209.09858}, 2022.

\bibitem[Drummond and Shearer(2006)]{drummond2006open}
N.~Drummond and R.~Shearer.
\newblock The open world assumption.
\newblock In \emph{eSI Workshop: The Closed World of Databases meets the Open World of the Semantic Web}, volume~15, page~1, 2006.

\bibitem[He et~al.(2024{\natexlab{a}})He, Noci, Paliotta, Schlag, and Hofmann]{he2024understanding}
B.~He, L.~Noci, D.~Paliotta, I.~Schlag, and T.~Hofmann.
\newblock Understanding and minimising outlier features in neural network training.
\newblock \emph{arXiv preprint arXiv:2405.19279}, 2024{\natexlab{a}}.

\bibitem[He et~al.(2016)He, Zhang, Ren, and Sun]{he2016deep}
K.~He, X.~Zhang, S.~Ren, and J.~Sun.
\newblock Deep residual learning for image recognition.
\newblock In \emph{Proceedings of the IEEE conference on computer vision and pattern recognition}, pages 770--778, 2016.

\bibitem[He et~al.(2024{\natexlab{b}})He, Yuan, Han, Wang, Su, Yin, Liu, and Gong]{he2024exploring}
R.~He, Y.~Yuan, Z.~Han, F.~Wang, W.~Su, Y.~Yin, T.~Liu, and Y.~Gong.
\newblock Exploring channel-aware typical features for out-of-distribution detection.
\newblock In \emph{Proceedings of the AAAI conference on artificial intelligence}, volume~38, pages 12402--12410, 2024{\natexlab{b}}.

\bibitem[Hendrycks and Gimpel(2016)]{hendrycks2016baseline}
D.~Hendrycks and K.~Gimpel.
\newblock A baseline for detecting misclassified and out-of-distribution examples in neural networks.
\newblock \emph{arXiv preprint arXiv:1610.02136}, 2016.

\bibitem[Hendrycks et~al.(2019)Hendrycks, Basart, Mazeika, Zou, Kwon, Mostajabi, Steinhardt, and Song]{hendrycks2019scaling}
D.~Hendrycks, S.~Basart, M.~Mazeika, A.~Zou, J.~Kwon, M.~Mostajabi, J.~Steinhardt, and D.~Song.
\newblock Scaling out-of-distribution detection for real-world settings.
\newblock \emph{arXiv preprint arXiv:1911.11132}, 2019.

\bibitem[Hsu et~al.(2020)Hsu, Shen, Jin, and Kira]{hsu2020generalized}
Y.-C. Hsu, Y.~Shen, H.~Jin, and Z.~Kira.
\newblock Generalized odin: Detecting out-of-distribution image without learning from out-of-distribution data.
\newblock In \emph{Proceedings of the IEEE/CVF conference on computer vision and pattern recognition}, pages 10951--10960, 2020.

\bibitem[Huang et~al.(2021)Huang, Geng, and Li]{huang2021importance}
R.~Huang, A.~Geng, and Y.~Li.
\newblock On the importance of gradients for detecting distributional shifts in the wild.
\newblock \emph{Advances in Neural Information Processing Systems}, 34:\penalty0 677--689, 2021.

\bibitem[Huang et~al.(2020)Huang, Kroening, Ruan, Sharp, Sun, Thamo, Wu, and Yi]{huang2020survey}
X.~Huang, D.~Kroening, W.~Ruan, J.~Sharp, Y.~Sun, E.~Thamo, M.~Wu, and X.~Yi.
\newblock A survey of safety and trustworthiness of deep neural networks: Verification, testing, adversarial attack and defence, and interpretability.
\newblock \emph{Computer Science Review}, 37:\penalty0 100270, 2020.

\bibitem[Krizhevsky et~al.(2009)Krizhevsky, Hinton, et~al.]{krizhevsky2009learning}
A.~Krizhevsky, G.~Hinton, et~al.
\newblock Learning multiple layers of features from tiny images.
\newblock 2009.

\bibitem[Lee et~al.(2018)Lee, Lee, Lee, and Shin]{lee2018simple}
K.~Lee, K.~Lee, H.~Lee, and J.~Shin.
\newblock A simple unified framework for detecting out-of-distribution samples and adversarial attacks.
\newblock \emph{Advances in neural information processing systems}, 31, 2018.

\bibitem[Liang et~al.(2017)Liang, Li, and Srikant]{liang2017enhancing}
S.~Liang, Y.~Li, and R.~Srikant.
\newblock Enhancing the reliability of out-of-distribution image detection in neural networks.
\newblock \emph{arXiv preprint arXiv:1706.02690}, 2017.

\bibitem[Litjens et~al.(2017)Litjens, Kooi, Bejnordi, Setio, Ciompi, Ghafoorian, Van Der~Laak, Van~Ginneken, and S{\'a}nchez]{litjens2017survey}
G.~Litjens, T.~Kooi, B.~E. Bejnordi, A.~A.~A. Setio, F.~Ciompi, M.~Ghafoorian, J.~A. Van Der~Laak, B.~Van~Ginneken, and C.~I. S{\'a}nchez.
\newblock A survey on deep learning in medical image analysis.
\newblock \emph{Medical image analysis}, 42:\penalty0 60--88, 2017.

\bibitem[Liu et~al.(2020)Liu, Wang, Owens, and Li]{liu2020energy}
W.~Liu, X.~Wang, J.~Owens, and Y.~Li.
\newblock Energy-based out-of-distribution detection.
\newblock \emph{Advances in neural information processing systems}, 33:\penalty0 21464--21475, 2020.

\bibitem[Metropolis and Ulam(1949)]{metropolis1949monte}
N.~Metropolis and S.~Ulam.
\newblock The monte carlo method.
\newblock \emph{Journal of the American statistical association}, 44\penalty0 (247):\penalty0 335--341, 1949.

\bibitem[Mudholkar and Hutson(2000)]{mudholkar2000epsilon}
G.~S. Mudholkar and A.~D. Hutson.
\newblock The epsilon--skew--normal distribution for analyzing near-normal data.
\newblock \emph{Journal of statistical planning and inference}, 83\penalty0 (2):\penalty0 291--309, 2000.

\bibitem[Netzer et~al.(2011)Netzer, Wang, Coates, Bissacco, Wu, Ng, et~al.]{netzer2011reading}
Y.~Netzer, T.~Wang, A.~Coates, A.~Bissacco, B.~Wu, A.~Y. Ng, et~al.
\newblock Reading digits in natural images with unsupervised feature learning.
\newblock In \emph{NIPS workshop on deep learning and unsupervised feature learning}, volume 2011, page~4. Granada, 2011.

\bibitem[Nguyen et~al.(2015)Nguyen, Yosinski, and Clune]{nguyen2015deep}
A.~Nguyen, J.~Yosinski, and J.~Clune.
\newblock Deep neural networks are easily fooled: High confidence predictions for unrecognizable images.
\newblock In \emph{Proceedings of the IEEE conference on computer vision and pattern recognition}, pages 427--436, 2015.

\bibitem[Park et~al.(2023)Park, Jung, and Teoh]{park2023nearest}
J.~Park, Y.~G. Jung, and A.~B.~J. Teoh.
\newblock Nearest neighbor guidance for out-of-distribution detection.
\newblock In \emph{Proceedings of the IEEE/CVF International Conference on Computer Vision}, pages 1686--1695, 2023.

\bibitem[Sehwag et~al.(2021)Sehwag, Chiang, and Mittal]{sehwag2021ssd}
V.~Sehwag, M.~Chiang, and P.~Mittal.
\newblock Ssd: A unified framework for self-supervised outlier detection.
\newblock \emph{arXiv preprint arXiv:2103.12051}, 2021.

\bibitem[Song et~al.(2022)Song, Sebe, and Wang]{song2022rankfeat}
Y.~Song, N.~Sebe, and W.~Wang.
\newblock Rankfeat: Rank-1 feature removal for out-of-distribution detection.
\newblock \emph{Advances in Neural Information Processing Systems}, 35:\penalty0 17885--17898, 2022.

\bibitem[Sun and Li(2022)]{sun2022dice}
Y.~Sun and Y.~Li.
\newblock Dice: Leveraging sparsification for out-of-distribution detection.
\newblock In \emph{European Conference on Computer Vision}, pages 691--708. Springer, 2022.

\bibitem[Sun et~al.(2021)Sun, Guo, and Li]{sun2021react}
Y.~Sun, C.~Guo, and Y.~Li.
\newblock React: Out-of-distribution detection with rectified activations.
\newblock \emph{Advances in Neural Information Processing Systems}, 34:\penalty0 144--157, 2021.

\bibitem[Sun et~al.(2022)Sun, Ming, Zhu, and Li]{sun2022out}
Y.~Sun, Y.~Ming, X.~Zhu, and Y.~Li.
\newblock Out-of-distribution detection with deep nearest neighbors.
\newblock In \emph{International Conference on Machine Learning}, pages 20827--20840. PMLR, 2022.

\bibitem[Van~der Maaten and Hinton(2008)]{van2008visualizing}
L.~Van~der Maaten and G.~Hinton.
\newblock Visualizing data using t-sne.
\newblock \emph{Journal of machine learning research}, 9\penalty0 (11), 2008.

\bibitem[Van~Horn et~al.(2018)Van~Horn, Mac~Aodha, Song, Cui, Sun, Shepard, Adam, Perona, and Belongie]{van2018inaturalist}
G.~Van~Horn, O.~Mac~Aodha, Y.~Song, Y.~Cui, C.~Sun, A.~Shepard, H.~Adam, P.~Perona, and S.~Belongie.
\newblock The inaturalist species classification and detection dataset.
\newblock In \emph{Proceedings of the IEEE conference on computer vision and pattern recognition}, pages 8769--8778, 2018.

\bibitem[Wang et~al.(2022)Wang, Li, Feng, and Zhang]{wang2022vim}
H.~Wang, Z.~Li, L.~Feng, and W.~Zhang.
\newblock Vim: Out-of-distribution with virtual-logit matching.
\newblock In \emph{Proceedings of the IEEE/CVF conference on computer vision and pattern recognition}, pages 4921--4930, 2022.

\bibitem[Xiao et~al.(2010)Xiao, Hays, Ehinger, Oliva, and Torralba]{xiao2010sun}
J.~Xiao, J.~Hays, K.~A. Ehinger, A.~Oliva, and A.~Torralba.
\newblock Sun database: Large-scale scene recognition from abbey to zoo.
\newblock In \emph{2010 IEEE computer society conference on computer vision and pattern recognition}, pages 3485--3492. IEEE, 2010.

\bibitem[Xu et~al.(2015)Xu, Ehinger, Zhang, Finkelstein, Kulkarni, and Xiao]{xu2015turkergaze}
P.~Xu, K.~A. Ehinger, Y.~Zhang, A.~Finkelstein, S.~R. Kulkarni, and J.~Xiao.
\newblock Turkergaze: Crowdsourcing saliency with webcam based eye tracking.
\newblock \emph{arXiv preprint arXiv:1504.06755}, 2015.

\bibitem[Yang et~al.(2021)Yang, Zhou, Li, and Liu]{yang2021generalized}
J.~Yang, K.~Zhou, Y.~Li, and Z.~Liu.
\newblock Generalized out-of-distribution detection: A survey.
\newblock \emph{arXiv preprint arXiv:2110.11334}, 2021.

\bibitem[Yang et~al.(2022)Yang, Wang, Zou, Zhou, Ding, Peng, Wang, Chen, Li, Sun, et~al.]{yang2022openood}
J.~Yang, P.~Wang, D.~Zou, Z.~Zhou, K.~Ding, W.~Peng, H.~Wang, G.~Chen, B.~Li, Y.~Sun, et~al.
\newblock Openood: Benchmarking generalized out-of-distribution detection.
\newblock \emph{Advances in Neural Information Processing Systems}, 35:\penalty0 32598--32611, 2022.

\bibitem[Yu et~al.(2015)Yu, Seff, Zhang, Song, Funkhouser, and Xiao]{yu2015lsun}
F.~Yu, A.~Seff, Y.~Zhang, S.~Song, T.~Funkhouser, and J.~Xiao.
\newblock Lsun: Construction of a large-scale image dataset using deep learning with humans in the loop.
\newblock \emph{arXiv preprint arXiv:1506.03365}, 2015.

\bibitem[Zhang and Xiang(2023)]{zhang2023decoupling}
Z.~Zhang and X.~Xiang.
\newblock Decoupling maxlogit for out-of-distribution detection.
\newblock In \emph{Proceedings of the IEEE/CVF Conference on Computer Vision and Pattern Recognition}, pages 3388--3397, 2023.

\bibitem[Zhou et~al.(2017)Zhou, Lapedriza, Khosla, Oliva, and Torralba]{zhou2017places}
B.~Zhou, A.~Lapedriza, A.~Khosla, A.~Oliva, and A.~Torralba.
\newblock Places: A 10 million image database for scene recognition.
\newblock \emph{IEEE transactions on pattern analysis and machine intelligence}, 40\penalty0 (6):\penalty0 1452--1464, 2017.

\bibitem[Zhu et~al.(2022)Zhu, Chen, Xie, Li, Zhang, Xue, Tian, Chen, et~al.]{zhu2022boosting}
Y.~Zhu, Y.~Chen, C.~Xie, X.~Li, R.~Zhang, H.~Xue, X.~Tian, Y.~Chen, et~al.
\newblock Boosting out-of-distribution detection with typical features.
\newblock \emph{Advances in Neural Information Processing Systems}, 35:\penalty0 20758--20769, 2022.

\end{thebibliography}

\end{document}